\documentclass[conference]{IEEEtran}
\IEEEoverridecommandlockouts
\usepackage{cite}
\usepackage{amsmath,amssymb,amsfonts}
\usepackage{algorithmic}
\usepackage{graphicx}
\usepackage{multirow}
\usepackage{textcomp}
\usepackage{enumitem}
\usepackage{xcolor}
\usepackage{balance}
\def\BibTeX{{\rm B\kern-.05em{\sc i\kern-.025em b}\kern-.08em
    T\kern-.1667em\lower.7ex\hbox{E}\kern-.125emX}}
\begin{document}

\title{ProtoConNet: Prototypical Augmentation and  Alignment for Open-Set Few-Shot Image Classification\\
}
\author{Kexuan Shi\IEEEauthorrefmark{1}, Zhuang Qi\IEEEauthorrefmark{1}, Jingjing Zhu, Lei Meng\IEEEauthorrefmark{2}, Yaochen Zhang, Haibei Huang, Xiangxu Meng}

\maketitle

\begin{abstract}
Open-set few-shot image classification aims to train models using a small amount of labeled data, enabling them to achieve good generalization when confronted with unknown environments. Existing methods mainly use visual information from a single image to learn class representations to distinguish known from unknown categories. However, these methods often overlook the benefits of integrating rich contextual information. To address this issue, this paper proposes a prototypical augmentation and alignment method, termed ProtoConNet, which incorporates background information from different samples to enhance the diversity of the feature space, breaking the spurious associations between context and image subjects in few-shot scenarios. Specifically, it consists of three main modules: the clustering-based data selection (CDS) module mines diverse data patterns while preserving core features; the contextual-enhanced semantic refinement (CSR) module builds a context dictionary to integrate into image representations, which boosts the model's robustness in various scenarios; and the prototypical alignment (PA) module reduces the gap between image representations and class prototypes, amplifying feature distances for known and unknown classes. Experimental results
from two datasets verified that ProtoConNet enhances the effectiveness of representation learning in few-shot scenarios and identifies open-set samples, making it superior to existing methods. 
\end{abstract}

\begin{IEEEkeywords}
Open-set recognition, few-shot learning, feature augmentation, Jittor framework.
\end{IEEEkeywords}

\section{Introduction}
Open-set few-shot visual classification, which aims to leverage a limited amount of data for effective recognition of both known and unknown categories, has attracted widespread attention across various fields \cite{pal2024few,su2024toward,yin2024rd,peng2022few}. It addresses the challenges of data scarcity and high labeling costs \cite{che2023boosting,huang2024feature}. Recent work typically involves fine-tuning models that have been pre-trained on large datasets, leveraging their rich feature representations to adapt to new data \cite{zhou2022learning,zhou2022conditional,su2024toward,park2024pre}. However, existing studies indicate that the limited number of samples for known classes may result in inadequate generalization ability of the model for these classes and the fine-tuning process could lead to knowledge forgetting, which in turn results in decreased performance on unknown categories \cite{khattak2023self,khattak2023maple,li2024learning,tian2024survey}. 

\begin{figure}[t]
  \centering
  \includegraphics[width=0.951\linewidth]{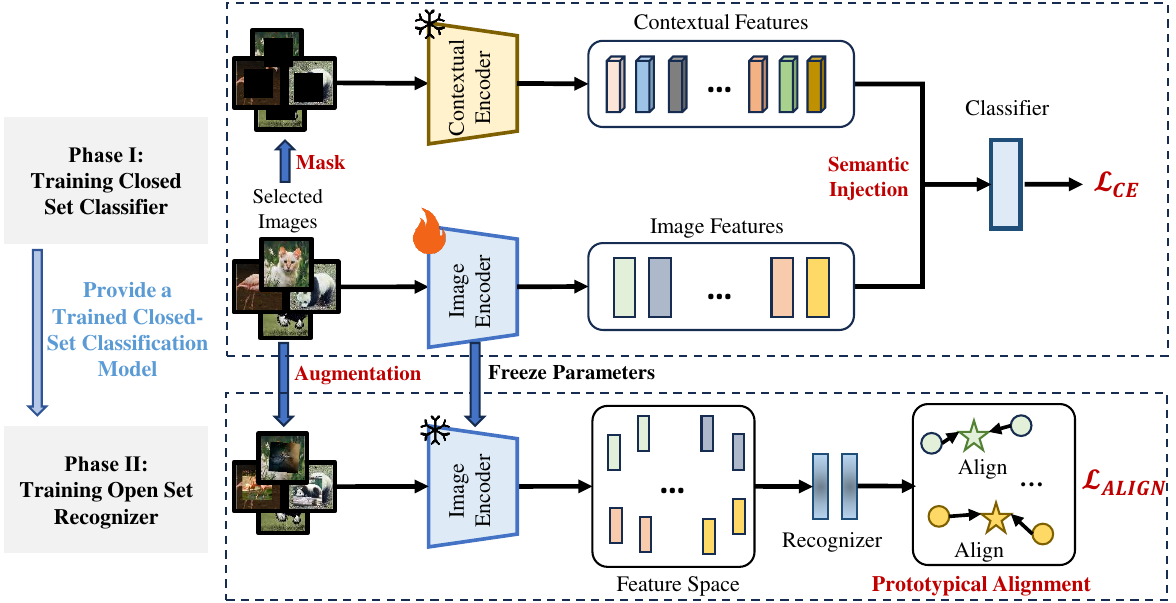}
  \caption{Illustration of the ProtoConNet workflow. It enhances the diversity of the feature space by integrating contextual features, which improves the generalization ability of the closed-set classification model. Secondly, it performs prototype alignment to reduce the gap between prototypes and known class features, resulting in a robust open set recognizer.}
  \label{fig1}
\end{figure} 

To improve the performance of models in open-set few-shot scenarios, current approaches typically focus on two main strategies. On the one hand, data augmentation methods aim to enhance sample diversity by adjusting data distribution or expanding the dataset \cite{Pal_2023_WACV,wang2024enhance,shao2024collaborative}. However, these techniques often come with additional computing costs, such as training GANs or using diffusion models to generate samples \cite{li2024joint}, and they may exacerbate the problem of knowledge forgetting when dealing with unseen classes. For example, the fine-tuning of the BEiT model \cite{peng2022beit} focuses on optimizing seen categories, making it difficult to provide effective recognition for unseen categories. On the other hand, fine-tuning multimodal pre-trained models leverages knowledge from both images and text for collaborative decision-making. These methods often design parameter-efficient fine-tuning strategies to maintain generalization ability \cite{zhou2022learning,khattak2023self,zhu2024croft}. Nevertheless, their performance on known classes is usually inferior to that of methods specifically designed for closed-set recognition. Therefore, new approaches are urgently needed to balance the recognition of known classes with the generalization to unseen classes.

To address this problem, this paper proposes a feature augmentation and prototypical alignment method, termed ProtoConNet, which designs an open-set recognizer to establish connections between multiple models, enabling the use of specific models to handle open-set and closed-set samples. Specifically, ProtoConNet comprises three core components: the clustering-based data selection (CDS) module, the contextual-enhanced semantic  refinement (CSR) module, and the prototypical alignment (PA) module. To avoid the uncertainty caused by randomly selecting samples, the CDS module employs a clustering method to mine data patterns, which preserves both the core features and diversity of the samples. The CSR module leverages the diverse samples selected by the CDS module to extract rich contextual information and integrates it into image features, which breaks the spurious associations between the subject and background of the samples. Subsequently, the PA module utilizes the open-set recognizer to align image representations with the corresponding class prototypes, reducing the differences between them while simultaneously amplifying the distance between features of unknown and known classes, which enables ProtoConNet determine the decision-making path to fully leverage the strengths of different models.

Extensive experiments were conducted on two datasets, as well as performance comparison, ablation studies of the three components, and case studies of key components. Experimental results demonstrate that ProtoConNet improves the model's attention to image subjects in few-shot scenarios and effectively distinguishes between known and unknown classes. In conclusion, this study has three main contributions as follows:

\begin{itemize}[leftmargin=10pt]
    \item 
   This study proposes a model-agnostic framework (ProtoConNet) for open-set few-shot learning based on Jittor framework, which can be integrated as a plug-and-play component into any backbone network. 
    \item 
    Experiments have found that few-shot learning often reduces the model's attention to image subjects in unseen environments. And integrating contextual information from diverse samples can mitigate this issue, thereby enhancing the model's generalization ability.
    \item 
    
    The entire experimental code is implemented using the Jittor framework, and we have contributed to the development of the Jittor platform by adding several custom features, including Grad-CAM and IVLP models in Jittor versions.
\end{itemize}
\section{Related Work}

\subsection{Few-Shot Image Classification}

Few-Shot Learning (FSL) aims to develop accurate models with limited samples, primarily through data augmentation and meta-learning. Data augmentation methods, like DiffusionCLS \cite{chen2024effective} and Diff-Mix \cite{wang2024enhance}, leverage pretrained diffusion models to diversify samples, while GAN-based approaches, such as MORGAN \cite{Pal_2023_WACV} and DAIC-GAN \cite{astolfi2023instance}, generate synthetic data. However, these often require extra models and data, conflicting with Jittor AI Challenge rules. Meta-learning methods seek generalizable initializations, as seen in LEO's low-dimensional optimization \cite{rusu2019meta}. Although integrating pretrained models like DeiT \cite{touvron2022deit}, MAE \cite{he2022masked}, and BEiT \cite{peng2022beit} has led to notable results, challenges with unknown classes remain

\subsection{Open-Set Recognition}

Open set recognition (OSR) aims to classify known classes precisely while identifying unknown classes. Traditional methods, like R3CBAM, use specialized architectures (e.g., CBAM3D layers) to enhance relevant features and reject outliers without needing to set thresholds \cite{pal2022fewshot}. However, they often only detect unknown classes without fine-grained classification. Recent approaches fine-tune pretrained multimodal models for OSR, balancing generalization to unknown classes with optimizing known class samples. Parameter-efficient methods like CoOp \cite{zhou2022learning} and IVLP \cite{khattak2023maple} support this, though they may underperform in recognizing known classes compared to closed set recognition models.

\subsection{Few-Shot Open-Set Recognition}
Few-shot open set recognition is a relatively under-explored field, where researchers typically combine OSR with FSL ideas to tackle OSR issues with limited training samples. For example, MRM uses a hypersphere with a learnable radius to separate classes and captures class distribution by evaluating sample similarities~\cite{che2023mrm}. SnaTCHer introduces task-adaptive transformation functions into FSL methods and quantifies transformation differences to reject unknown samples~\cite{jeong2021snatcher}. REFOCS proposes using image reconstruction and enhanced latent representations to detect out-of-distribution samples~\cite{nag2023refocs}.  GEL introduces class- and pixel-wise similarity modules along with an energy-based module that assigns high energy to unknown classes, improving the classifier's ability to reject samples belong to unknown classes~\cite{wang2023gel}.

\section{Problem Formulation}

This paper is set against the backdrop of the \textbf{Jittor AI Challenge }\footnote{https://www.educoder.net/competitions/index/Jittor-5} and follows the competition's framework. Suppose there is an image dataset containing $C$ classes, denoted as $\{1,...,C\}$. For class $c$, there are $N_c$ samples, i.e. $\{x_1,...,x_{N_c}\}$. In this task, \textbf{we need to manually select 4 samples from each class to assist in model training}, which simulates a few-shot visual classification case. There are two objectives: \textbf{the TestA dataset contains only samples from classes that appeared in the training set, while the TestB dataset also includes samples from several classes that did not appear in the training set}. Baseline methods typically randomly select samples to train the model and optimize the classification loss, i.e. $\mathcal{L}_{cls}=CE(F(x_{select}),\hat{y})$, where $x_{select}$ is a selected sample and $f(\cdot)$ is a classification model, $\hat{y}$ denotes the ground-truth of  $x_{select}$. 

\begin{figure*}[t]
  \centering
  \includegraphics[width=0.951\linewidth]{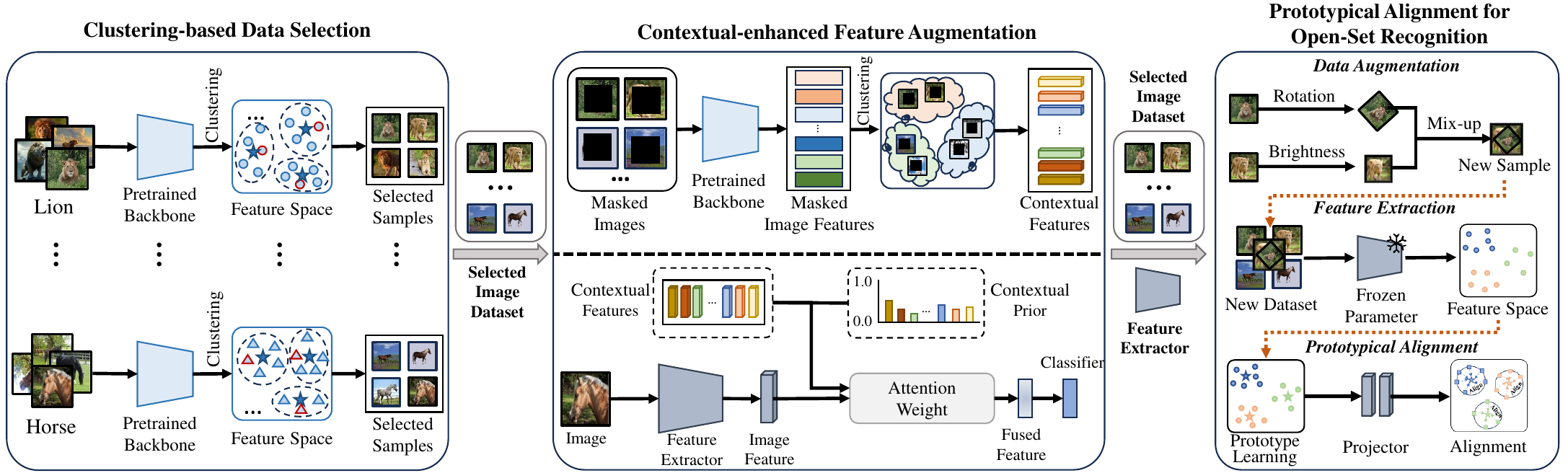}
  \caption{Illustration of the training framework of ProtoConNet. It contains three main modules, where the clustering-based data selection module utilizes a pretrained backbone to extract image features and performs clustering to mine different data patterns, which provides the selected dataset for subsequent modules. The contextual-enhanced semantic  refinement module generates various contextual features to fuse into image features to enhance feature diversity. The prototypical alignment for open-set recognition module generates new samples by random transformation and combination and learns a projector to align image features and the corresponding prototype, which can recognize samples of unknown categories.}
  \label{fig2}
\end{figure*}

In contrast, the proposed ProtoConNet introduces improvements in three aspects. Firstly, ProtoConNet design a clustering-based data selection method, which can mine different distribution patterns of samples with the same label, i.e., $\{clu_{1}^i,...,clu_{k}^i\}=kmeans(f_{i},k)$, where $kmean(\cdot)$ is a clustering method, $f_{i}$ denotes image features of class $i$, $k$ is a hyperparameter, which indicates the number of selected samples. And $\{x_1^i,...,x_k^i\}$ represent the samples closest to the center of each of clusters $\{clu_{1}^i,...,clu_{4}^i\}$, respectively. Secondly, we develop a cross-class contextual augmentation method to generate masks, which occludes the main object in the image and retains contextual information, i.e., $x'=mask(x)$, where $x$ and $x'$ represent the original and masked images, respectively. Each sample feature $f_x$ can incorporate arbitrary contextual features $f_{other}$ from other classes to enhance the diversity of its corresponding features, i.e., $f_{fuse}=f_x\oplus f_{other}$. Thirdly, we design a open-set recognition method, which learns a mapping $\mathcal{M}(\cdot)$ to align image features $\{f_i\}$ and the corresponding prototypes $\{p_1,...,p_{C}\}$, i.e. $f_i\xrightarrow{\mathcal{M}(\cdot)}p_i$. In the inference phase, $f_i'=\mathcal{M}(F(x_i'))$, and $s=cossim(f_i',p_i)$ denotes a similarity score. If $s>T$, it indicates that the sample belongs to an unknown class; otherwise, it belongs to a known class.

\section{Method}

\subsection{Framework}

This paper proposes a contextual-enhanced few-shot open-set recognition method, termed ProtoConNet, which improves the diversity of feature space to avoid the negative impact of insufficient samples. As shown in Figure \ref{fig2}, ProtoConNet comprises three main modules: the clustering-based data selection (CDS) module, which uses clustering to select representative samples and reduce instability; the contextual-enhanced semantic  refinement (CSR) module, which incorporates contextual information to enhance feature diversity and robustness; and the prototypical alignment (PA) module, which aligns image features with prototypes to distinguish unseen class samples from training samples.

\subsection{Clustering-based Data Selection}
The Cluster-based Data Selection (CDS) module aims to select representative samples from multiple categories to ensure that these samples capture the key features of the data distribution. Specifically, it employs clustering methods to group the samples and identifies those that contain diverse and informative features. Firstly, it extracts image features by using a pre-trained backbone (such as CLIP \cite{clip}), which can be expressed by
\begin{equation}
    f = CLIP(x),
\end{equation}
where $f$ is the image feature of the sample $x$. To fully explore data patterns, the CDS module employs a kmeans method \cite{qi2023cross} to group similar samples into the same cluster, i.e.,
\begin{equation}
    c_i^1,...,c_i^k = KMEANS(f_i, k),
\end{equation}
where $f_i$ represents a feature set with label $i$, $k$ denotes the number of clusters, and $c_i^k$ is the k-th cluster of class $i$. $KMEAN(\cdot)$ is a clustering method that initializes cluster centers randomly and iteratively refines them. Then, the CDS module selects representative samples as those closest to their cluster centroids based on Euclidean distance, i,e.,
\begin{equation}
    x_i^k = min\{dist(f_i^k, center_i)|f_i^k \in c_i^k\},
\end{equation}
where $f_i^k$ and $center_i$ represent a feature and the cluster center in cluster $c_i^k$, while $x_i^k$ is the sample closest to the centroid. This approach benefits few-shot classification by selecting the most representative samples from each category, reducing dataset size while maintaining diversity, and enhancing the model's generalization under limited data.

\subsection{contextual-enhanced semantic  refinement}
In few-shot scenarios, data scarcity limits the model's generalization ability. To address this, we propose a contextual-enhanced semantic  refinement (CSR) strategy that refines data representations by utilizing contextual cues from both the target object and its surroundings. Specifically, the CSR module masks the target region in the input image, highlighting the surrounding context to help the model learn background or environment-related features, enhancing generalization to unseen samples, 
\begin{equation}
    x_m = MASK(x, \gamma ),
\end{equation}
where $MASK(\cdot, \gamma)$ represents a zero mask of size $\gamma$ generated around the center of the image. $x_m$ is the masked image. Secondly, the CSR module uses pre-trained backbone to extract contextual features $f_m$ i.e.,
\begin{equation}
    f_m = Backbone(x_m),
\end{equation}
Subsequently, we use KMEANS method to divide contextual features into $\beta$ groups, i.e.,
\begin{equation}
    g_1,...,g_{\beta}=KMEANS(\{f_m\}, \beta),
\end{equation}
where $g_{\beta}$ is the $\beta$-th group. Then, the CSR module learns a set of contextual prototypes ${Z} = [{z}_1, {z}_2, ..., {z}_{\beta}]$ by averaging the intra-group contextual features, i.e.,
\begin{equation}
    z_i = \frac{1}{|g_i|} {\textstyle \sum_{j=1}^{|g_i|}}  f_m^j, f_m^j\in g_i,
\end{equation}
where $|g_i|$ is the size of group $g_i$, and $z_i$ is the corresponding contextual prototype. Notably, feature averaging reduces interference from partially covered targets. To enhance feature diversity with limited data, the CSR module incorporates contextual information from multiple perspectives into image features. This improves the model's ability to capture contextual relationships, boosting accuracy and robustness even with limited training samples, as expressed by:
\begin{equation}
    f_{fuse} = f +  {\textstyle \sum_{i=1}^{\beta}} \lambda_i z_i P(z_i) 
\end{equation}
where $f=F(x)$ represents the image feature extracted by $F(\cdot)$, $\lambda_i$ is a weight defined as $\lambda_{i}=\operatorname{softmax}\left(\frac{\left(\boldsymbol{W}{q} \boldsymbol{f}\right)^{T}\left(\boldsymbol{W}_{k} \boldsymbol{z}_{i}\right)}{\sqrt{d}}\right)$, where $d$ is the dimension of the contextual prototype and $\boldsymbol{W}_{q}$ and $\boldsymbol{W}_{k}$ are learnable weight matrices. The probability $P(z_i)$ is given by $P(z_i)=\frac{|g_i|}{\sum{i=1}^{\beta}|g_i|}$. A classifier $H(\cdot)$ is then used to make predictions, and the Cross-Entropy loss optimizes both the feature extractor and the classifier, expressed as:
\begin{equation}
    \mathcal{L}_{CE}=CE(\hat{y},y),
\end{equation}
where $\hat{y}=H(f_{fuse})$ and $y$ are the prediction and the ground-truth of sample $x$, respectively.

\subsection{Prototypical Alignment for Open-Set Recognition}
To classify both known and unknown classes, the prototypical alignment (PA) module incorporates an open-set recognizer to enhance feature representation by aligning features with class prototypes. This improves recognition of known categories while distinguishing unseen ones. The module involves three key steps: data augmentation, feature extraction, and prototypical alignment. Since limited samples often lead to overfitting, the PA module addresses this by generating new samples for each class through transformations and mixup of transformed samples, i.e.,
\begin{equation}
    x_{new} = mixup(trans(A), trans(B)),
\end{equation}
where $trans(\cdot)$ is a random transformation function, such as rotation,  flipping, and scaling. $A$ and $B$ are different samples with the same label, and $x_{new}$ shares label with sample $A$ and $B$. $mixup(\cdot)$ is a data augmentation method \cite{qi2024cross}. This method enhances sample diversity, providing sufficient information for training open-set recognizer.

Furthermore, the PA module leverages feature extractor $F(\cdot)$ learned from the CSR module to extract image features $\{f_{pre}\}$ from original dataset $D$ and new samples $D_{new}=\{x_{new}\}$, i.e.,
\begin{equation}
    f_{pre} = F(x),\quad x \in D \quad or \quad x\in D_{new},
\end{equation}
And the class prototypes $[u_1,...,u_C]$ can be obtained by
\begin{equation}
    u_i =  \frac{1}{|f_{pre}^i|} {\textstyle \sum_{f_j\in f_{pre}^i} f_j} ,
\end{equation}
where $f_{pre}^i$ denotes image features set with label $i$.

\begin{figure}[t]
  \centering
  \includegraphics[width=0.9\linewidth]{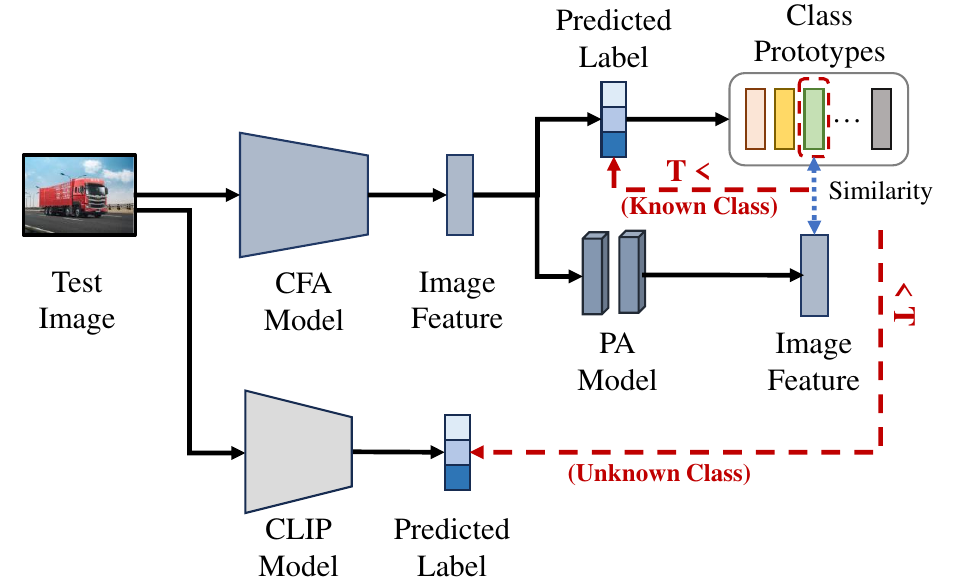}
  \caption{Illustration of the ProtoConNet inference workflow. The CSR model predicts a label to select a class prototype, which is compared to image features from the PA model. If the similarity exceeds threshold T, the CSR model provides the final prediction; otherwise, the CLIP model does.}
  \label{fig3}
\end{figure}

To obtain an open set recognizer based on only known class samples, the PA module aligns the image feature and the corresponding class prototypes. And we use Mean Square Error ($MSE(\cdot)$) to minimize the feature differences between training samples and their class prototypes, i.e.,
\begin{equation}
    \mathcal{L}_{ALIGN}=MSE(R(f_{pre}^i), u_i)),
\end{equation}
where $R(\cdot)$ is an open-set recognizer that helps identify known classes and detect outliers, critical for open-set recognition. As shown in Figure \ref{fig3}, ProtoConNet first uses the CSR model to generate initial predictions and selects class prototypes accordingly. It then calculates the similarity between image features from the PA model and the class prototypes. If the similarity exceeds threshold T, the sample is classified as a known class, with the CSR model providing the final prediction. Otherwise, it is classified as unknown, and the pre-trained CLIP model makes a zero-shot prediction.

\subsection{Training Strategy}

\begin{figure}[t]
  \centering
  \includegraphics[width=0.86\linewidth]{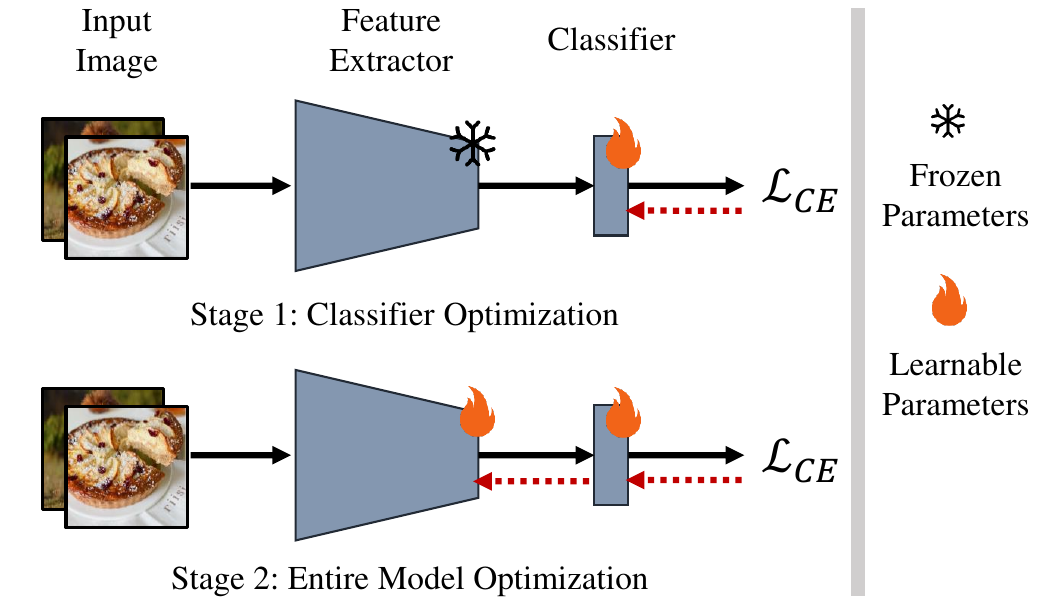}
  \caption{Illustration of the training strategy of the CSR module. It has a two-stage model optimization process. In Stage 1, only the classifier is optimized, while in Stage 2, the entire model is optimized. }
  \label{fig4}
\end{figure}

ProtoConNet uses the CSR module to train a closed-set classifier that can accurately identify labels for known class samples. The CSR module has the following loss function:
\begin{equation}
    \mathcal{L}_{CSR} = \mathcal{L}_{CE}.
\end{equation}
We found that directly optimizing both the pre-trained feature extractor and the randomly initialized classifier reduces performance. This happens because the feature extractor is already well-trained, while the classifier is still in its initial state. Using the same learning rate for both causes excessive updates to the feature extractor, which can harm its effectiveness. Additionally, the classifier struggles to use the features from the extractor in the early stages of training, causing instability. To address this, we designed a two-stage optimization process in the CSR module: first, optimize the classifier, then refine the entire model, as shown in Figure \ref{fig4}..

Subsequently, ProtoConNet uses the PA module to train an open-set recognizer that classifies samples as known or unknown. It has the following loss function:
\begin{equation}
    \mathcal{L}_{PA}=\mathcal{L}_{ALIGN}.
\end{equation}

\section{Experiments}

\subsection{Experimental Settings}
\vspace{-0.4cm}
\begin{table}[h]
\centering
\caption{The statistical information of the datasets.}
\vspace{-0.2cm}
\label{tab1}
\begin{tabular}{c|c|c}
\hline
\#Datasets            & \#Classes & \#Samples \\ \hline
Comprehensive Dataset & 374       & 185,872   \\ \hline
Tsinghua Dogs         & 130       & 70,428    \\ 
Caltech-101           & 91        & 9,144     \\ 
Food-101              & 101       & 101,000   \\ 
Animals               & 52        & 5,300     \\ \hline
\end{tabular}
\end{table}
\subsubsection{Datasets}
Following the \textbf{Jittor AI Challenge}\footnote{https://www.educoder.net/competitions/index/Jittor-5}, we use a comprehensive dataset to validate our method, including Tsinghua Dogs, Caltech-101, Food-101, and Animals, as detailed in Table \ref{tab1}. In the Jittor AI Challenge, duplicate categories from Caltech-101 and other datasets were removed, leaving 91 categories. \textbf{Participants can select 4 images per class for training, but no additional labeled or unlabeled data is allowed.} The competition has two leaderboards: Test A, with a closed-set test set, and Test B, with an open-set test set including 29 unseen categories from the Stanford-Cars dataset.

\subsubsection{Evaluation Measures}
We ues the \textit{Top-1 Accurancy} to measure the performance of each model, i.e.,
\begin{equation}
\text{Accuracy} = \frac{1}{N} \sum_{i=1}^{N} \mathbb{1}(\hat{y}_i = y_i),
\end{equation}
where $N$ is the total number of samples, $\hat{y}_i$ is the predicted class label for the $i$-th sample, $y_i$ is the true class label, and $\mathbb{1}(\cdot)$ is the indicator function which returns 1 if the expression inside is true and 0 otherwise.

\subsubsection{Hyper-parameter Settings}
In all experiments, the initial learning rate was $5 \times 10^{-5}$ for all methods, except for the separately trained classifier, which used $1 \times 10^{-4}$. A batch size of 32 was used, and the \textit{Adam optimizer} was applied with a weight decay of $1 \times 10^{-5}$ to prevent overfitting. The learning rate decayed by 0.1 after 8 epochs, with a total of 17 training epochs. This configuration was consistent across all methods. In the CDS module, $k$ was set to 4. In the CSR module, $\gamma$ was chosen from $\{64,96, 128\}$ and $\beta$ from $\{32,64\}$. We tuned the threshold $T$ from $\{0.0021,0.0022,0.0023,0.0024\}$.

\begin{table}[t]
\centering
\caption{Performance comparison of all methods on the comprehensive dataset. All methods were run by three trials, and the mean and standard deviation are reported.}
\label{tab2}
\setlength{\tabcolsep}{1.2mm}{
\begin{tabular}{c|cc|cc}
\hline
\multirow{2}{*}{Methods} & \multicolumn{2}{c|}{TestA} & \multicolumn{2}{c|}{TestB} \\\cline{2-5}
                         & Single          & Two        & Single          & Two        \\\hline
CLIP$_{Pre}$             & 69.02           & ---        & 70.94           & ---        \\
CLIP$_{FT}$              & 72.45$\pm$1.1   & ---        & 73.15$\pm$1.1   & ---        \\
ProtoConNet$_{CLIP}$     & 75.27           & ---        & 74.38           & ---        \\\hline
IVLP                     & 73.46$\pm$1.1   & ---        & 75.62$\pm$1.1   & ---        \\
ProtoConNet$_{IVLP}$     & 76.14           & ---        & 78.49           & ---        \\\hline
MAE                      & 57.42$\pm$1.1   & 59.39$\pm$1.1 & 41.28$\pm$1.1 & 42.89$\pm$1.1 \\
ProtoConNet$_{MAE}$      & 61.23           & 62.58      & 54.47           & 58.92      \\\hline
DeiT                     & 67.95$\pm$1.1   & 72.69$\pm$1.1 & 51.66$\pm$1.1 & 53.43$\pm$1.1 \\
ProtoConNet$_{DeiT}$     & 69.06           & 75.24      & 69.39           & 74.36      \\\hline
BEiT                     & 70.62$\pm$1.1   & 76.72$\pm$1.1 & 46.98$\pm$1.1 & 51.44$\pm$1.1 \\
ProtoConNet$_{BEiT}$     & 76.62           & \textbf{83.62} & 72.49        & \textbf{81.43} \\\hline
\end{tabular}}
\end{table}

\subsection{Performance Comparison}
Following the Jittor AI Challenge rules, we validate ProtoConNet by comparing it with six SOTA methods: pre-trained CLIP (CLIP$_{Pre}$) \cite{clip}, CLIP with full fine-tuning (CLIP$_{FT}$), MAE \cite{he2022masked}, DeiT \cite{touvron2022deit}, BEiT \cite{peng2022beit}, and IVLP \cite{khattak2023self}. Results are reported for both a single-stage (Single) and a two-stage (Two) training strategy. For methods without ProtoConNet, we report mean and variance across three random seeds; for methods with ProtoConNet, we report performance on specific samples. The results are summarized in Table \ref{tab2}.

\begin{itemize}[leftmargin=10pt]
    \item 
    ProtoConNet topped both Test A and Test B leaderboards by tackling sample scarcity in few-shot tasks. It learns accurate representations and detects unknown samples in open-set tasks via data selection and augmentation.
    \item 
     ProtoConNet is plug-and-play, compatible with various backbone networks, and significantly enhances performance on few-shot open-set tasks, demonstrating its model-agnostic nature.
    
    \item 
     The two-stage training strategy is superior to directly optimizing the entire model. This is reasonable because it is difficult for a randomly initialized classifier to align with a pre-trained backbone network.
     
    \item 
    \textbf{The proposed solution (ProtoConNet$_{BEiT}$) ranked in the top 10 on both the A and B leaderboards of the Jittor AI Challenge, with over 3,000 teams participating.} This demonstrates its effectiveness in few-shot open-set image classification tasks.
\end{itemize}

\subsection{Ablation Study}
\begin{table}[t]
\centering
\caption{Ablation study on the effectiveness of different components of ProtoConNet based on BEiT and CLIP.}
\vspace{0.2cm}
\label{tab3}
\begin{tabular}{c|c|c}
\hline
Methods       & TestA                & TestB                \\\hline
BEiT          & 79.45±1.2                     & 49.66±1.1                     \\
+CDS          & 82.18                     & 52.28                     \\
+CDS+CSR     &  83.62                   & 53.47                    \\
+CDS+CSR+PA & --- & 81.43  \\\hline
CLIP          & 72.45±1.1                     &  72.03±1.1                     \\
+CDS          & 73.60                    & 73.38                     \\
+CDS+CSR     & 75.27                     & 72.19                     \\
+CDS+CSR+PA & --- & 74.38 \\\hline
\end{tabular}
\end{table}
In this section, we evaluate the effects of the different components integrated into the ProtoConNet model by conducting ablation studies. The results are summarized in Table \ref{tab3}.

\begin{itemize}[leftmargin=10pt]

    \item 
    Relying on the BEiT model alone limits recognition to visible categories, resulting in poorer performance on Test B compared to Test A. Incorporating the CDS module enhances performance by selecting samples that maintain core features and diversity.
  
    \item 
    Integrating the CSR module further improves performance by utilizing rich contextual information, mitigating erroneous associations between objects and backgrounds often caused by overfitting in few-shot settings.
    \item 
    The PA module identifies open-set samples, enabling ProtoConNet to utilize the pre-trained CLIP model during inference, thereby enhancing performance on Test B.
    \item 
    Integrating CLIP with both the CDS and CSR modules results in lower performance on Test B compared to CDS alone, likely due to overfitting on known samples, which hinders recognition of unknown categories

\end{itemize}

\subsection{The Impact of Threshold Selection on Open-Set Recognition and Model Performance}
\begin{figure}[t]
  \centering
  \includegraphics[width=0.951\linewidth]{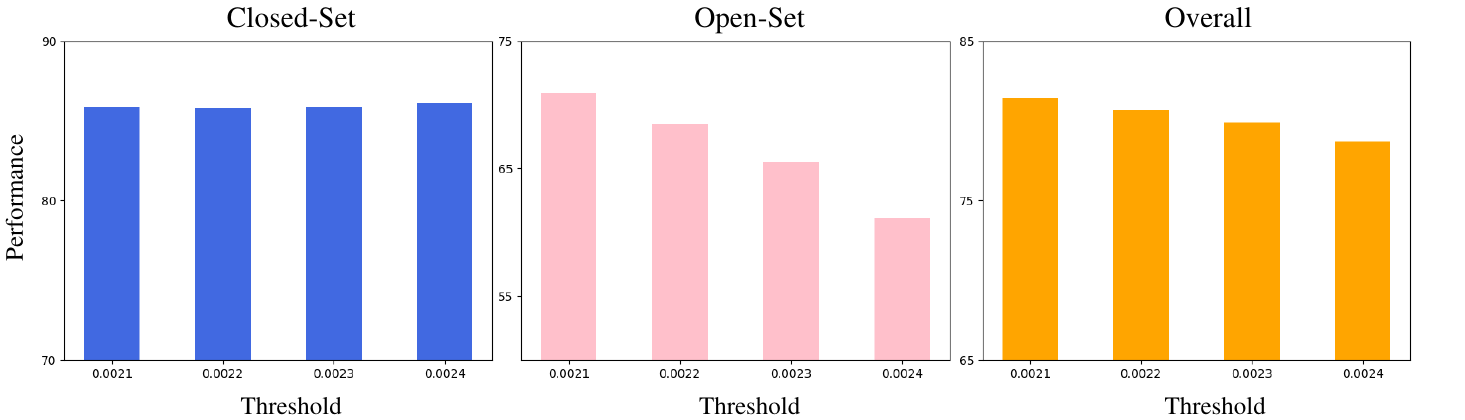}
  \caption{The impact of threshold selection on open-set recognition. We tuned the threshold $T$ from $\{0.0021,0.0022,0.0023,0.0024\}$.}
  \label{fig5}
\end{figure}

\begin{table}[h]
\centering
\caption{The performance of the model on open-set recognition affected by different thresholds.}
\label{tab4}
\begin{tabular}{ccccc}
\hline
Threshold & 0.0021 & 0.0022 & 0.0023 & 0.0024 \\\hline
Open-Set  & 86.03      & 82.66       & 78.73       & 74.71      \\\hline
\end{tabular}
\label{tab4}
\end{table}
This section investigates the impact of threshold selection on open-set recognition and model performance. As illustrated in Figure \ref{fig5}, we evaluated the model using four different threshold values and reported classification accuracy for closed-set, open-set, and overall datasets. \textbf{The results show that the model maintains strong stability in recognizing closed-set data, with minimal sensitivity to threshold variations}. However, for open-set data, recognition accuracy declines as the threshold increases, as shown in Table \ref{tab4}. This decline is primarily due to the model's increasing conservatism in identifying unknown categories, leading to misclassification of some unknown instances as known categories. This, in turn, results in a higher false negative rate for unseen samples, negatively impacting overall accuracy. Despite this, \textbf{the model’s performance on the overall dataset remains relatively stable, with only a slight decrease in accuracy as the threshold rises.}

\subsection{Case Studies}
\subsubsection{Clustering-based Data Selection}
\begin{figure}[h]
  \centering
  \includegraphics[width=0.95\linewidth]{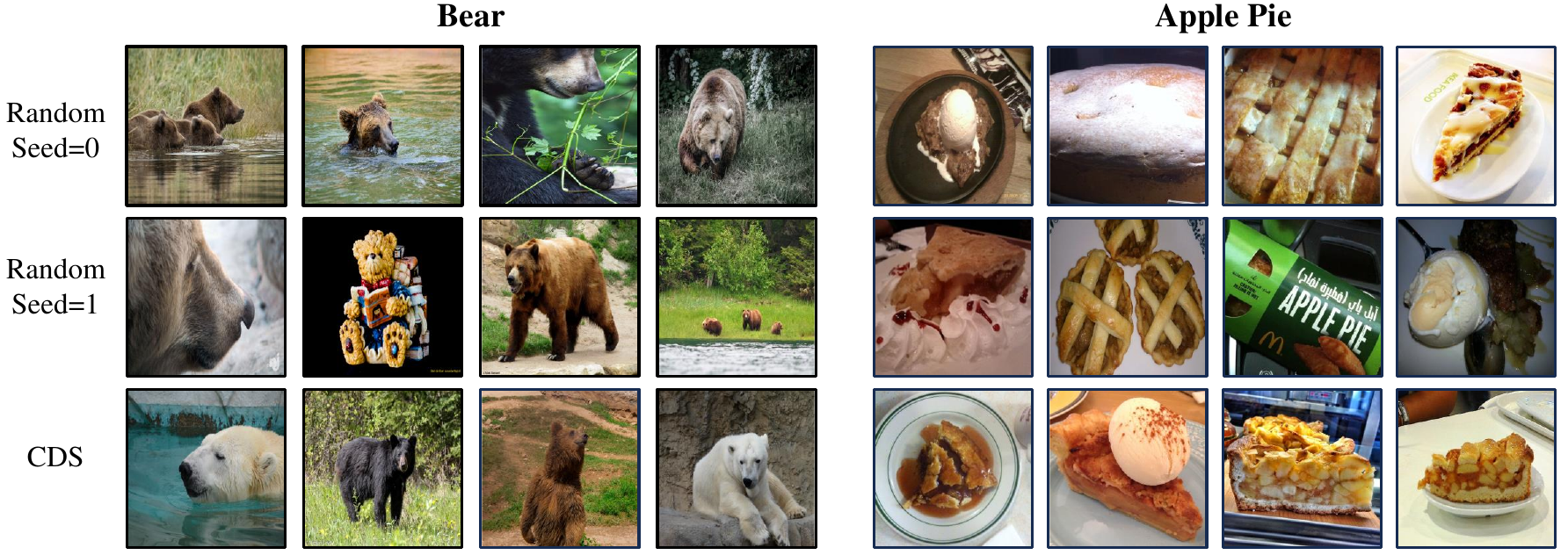}
  \caption{Comparison of results between the random sample selection method and the clustering-based data selection (CDS) method. The CDS method filters out poor samples with unclear core features and selects samples that encompass various forms of the subjects. }
  \label{fig:case1}
\end{figure}
This section evaluates the effectiveness of the Cluster-based Data Selection (CDS) method compared to random sampling. With limited sample sizes, random sampling often includes "poor" samples, such as those with less distinct or non-representative features, as shown in Figure \ref{fig:case1}. This worsens generalization in few-shot learning, as models trained on such data underperform. In contrast, \textbf{the CDS method identifies and excludes non-representative samples, reducing noise and improving training data quality.} By using clustering algorithms, CDS ensures dataset diversity, helping the model learn a broader range of features and improving generalization. CDS enhances model performance on unseen data, making it more applicable in real-world scenarios. Additionally, CDS can efficiently select key samples in other domains, outperforming random sampling and reducing the required training data.
\begin{figure}[h]
  \centering
  \includegraphics[width=0.9\linewidth]{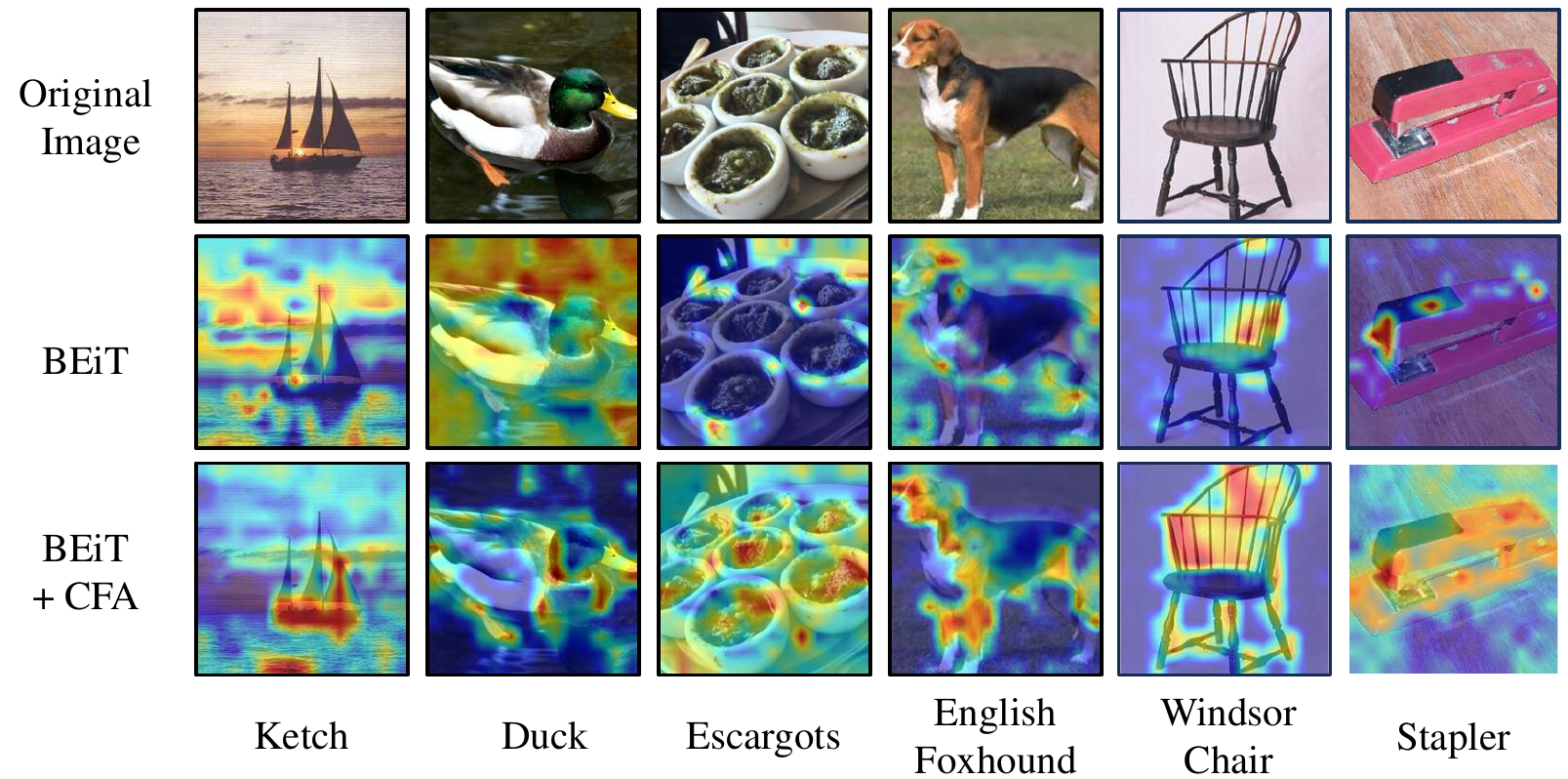}
  \caption{Visualization of attention on test samples for baseline methods and methods integrating the CSR module.}
  \label{fig7}
\end{figure}

\subsubsection{contextual-enhanced semantic  refinement}
This section uses Grad-CAM \cite{selvaraju2017grad} within the Jittor framework to explore the attention distribution of the baseline method and the CSR-integrated method on test samples. As shown in Figure \ref{fig7}, due to limited training data, the BEiT method learns suboptimal features, often focusing on the background rather than the main subject, as seen in the first, second, and fourth columns. To address this, the proposed CSR strategy combines original features with context prototypes in a weighted manner, enhancing feature representations. \textbf{The results show that CSR improves attention on the main subject, enabling more effective reasoning with new samples.} By incorporating contextual information, the model becomes less dependent on specific environments, achieving efficient feature learning even with limited data. For example, combining BEiT with CSR enhances focus on "Windsor Chair" and "Stapler.

\begin{figure}[h]
  \centering
  \includegraphics[width=0.9\linewidth]{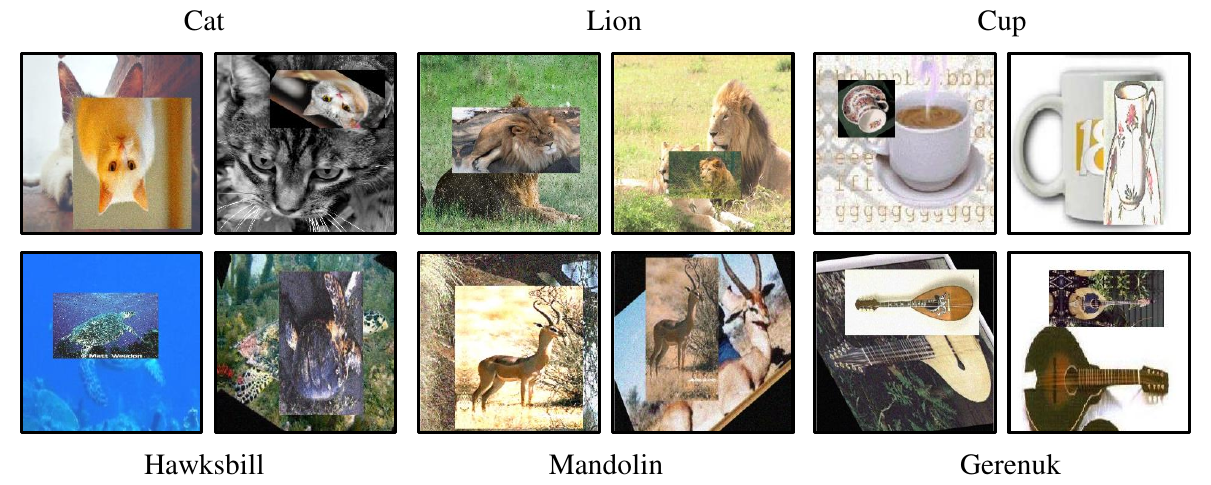}
  \caption{Showcase of generated samples. It retains the core features while increasing the diversity of the background.}
  \label{fig:case2}
\end{figure}

\subsubsection{Random Transformation and Combination for Data Augmentation}
This section evaluates the effectiveness of data augmentation methods in few-shot learning, focusing on Random Combination, Random Rotation and Flipping, and Noise Addition. As shown in Figure \ref{fig:case2}, augmented samples were generated using random rotations, flips, and noise addition, followed by random combinations to increase sample diversity. \textbf{The Paste Image Randomly method creates complex synthetic images by pasting small images onto larger backgrounds, enhancing data diversity and helping the model focus on object shapes and features instead of backgrounds.} \textbf{Random variations, such as rotation angles and noise types, further increase sample diversity.} The combination of these techniques significantly improves performance in few-shot learning tasks, enhancing model applicability in real-world scenarios.

\begin{figure}[h]
  \centering
  \includegraphics[width=0.9\linewidth]{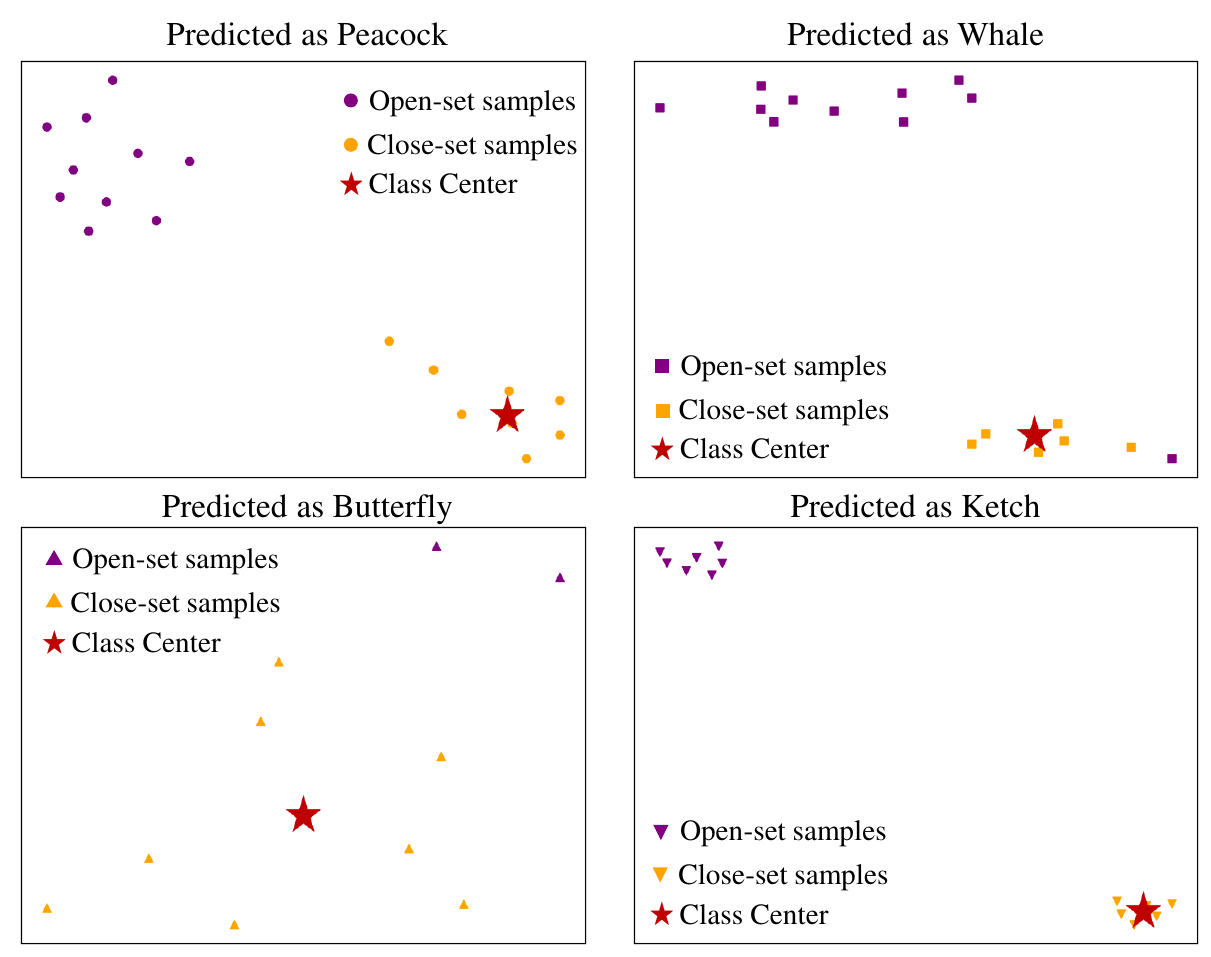}
  \caption{Visualization of feature distribution for samples predicted to belong to the same category shows that closed-set samples are closer to the class center, while open-set unknown samples are farther away.}
  \label{fig:case4}
\end{figure}
\subsubsection{Distance-based Open-Set Recognition}
This section highlights the importance of the prototypical alignment method in open-set recognition. During testing, unknown class samples are identified by comparing the distance between their features and class prototypes against a threshold. Using t-SNE \cite{van2008visualizing}, we visualized the features of samples predicted to belong to the same label. As shown in Figure \ref{fig:case4}, \textbf{closed-set samples are much closer to the class prototype than open-set samples, demonstrating how prototypical alignment enhances the feature gap between known and unknown classes.} For example, even if unknown samples are predicted as “Peacock,” their representation shows a clear gap from closed-set samples, facilitating class distinction.

\subsubsection{Error analysis}
\begin{figure}[h]
  \centering
  \includegraphics[width=0.99\linewidth]{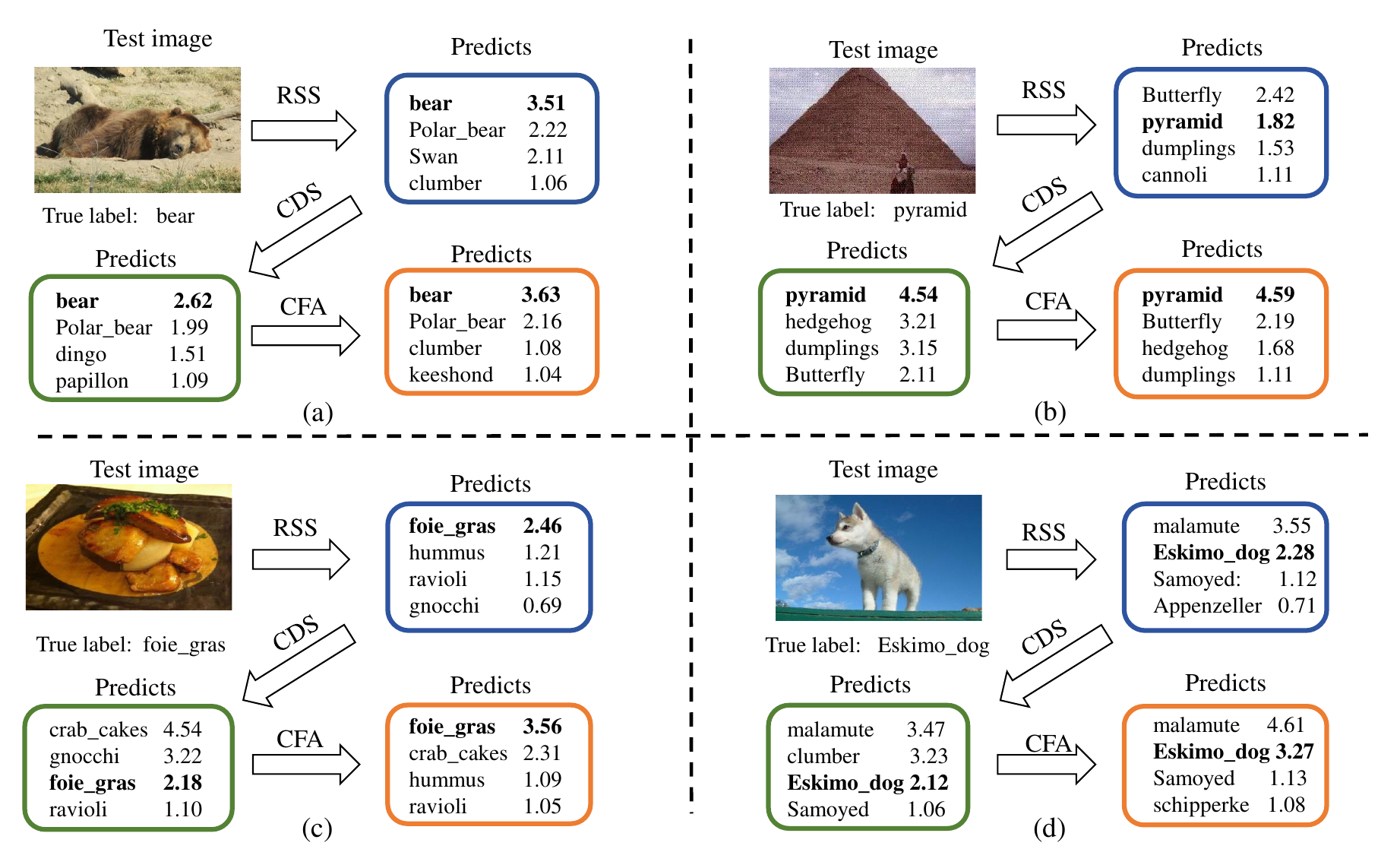}
  \caption{(a) The CDS and CSR modules enhanced the model's prediction for the Bear class. (b) The CDS module helped the model correct its errors. (c) The CSR module assisted the model in rectifying its mistakes. (d) The CSR module reduced the gap between the correct class and the top-1 prediction.}
  \label{fig10}
\end{figure}

This section presents an error analysis to highlight the role of the proposed modules, with RSS referring to the randomly selected samples method. As shown in Figure \ref{fig10}(a), both the CDS and CSR modules improve the model’s predictions. The CDS module selects more representative samples, while the CSR module enhances feature learning. Figure \ref{fig10}(b) shows that CDS helps the model correct errors, emphasizing the importance of sample selection. Figure \ref{fig10}(c) demonstrates that even when predictions fail, the CSR module can help correct errors by incorporating contextual information, improving generalization. In Figure \ref{fig10}(d), all methods made incorrect predictions, but the CSR module reduced the gap between the correct class and the top-1 prediction. These results confirm the effectiveness of the proposed methods.

\section{Conclusion}
This study proposes an open-set few-shot image classification framework (ProtoConNet) as part of the Jittor AI Challenge. The framework enhances model performance through three key modules, which address data selection, feature augmentation, and multi-model integration. Specifically, ProtoConNet utilizes the CDS method to select representative samples with rich feature diversity. The CSR module incorporates contextual information to reduce spurious associations between image subjects and backgrounds. The PA module leverages an open-set recognizer to establish connections between multiple models, optimizing their complementary strengths. The results indicate that ProtoConNet significantly improves the model’s focus on image subjects and effectively detects unknown samples introduced by open-set tasks.

Despite ProtoConNet's Top-10 results on both leaderboards, three areas remain for improvement. First, First, improving the open-set recognizer could support collaborative inference in federated learning. Second, combining feature augmentation with causal reasoning could enhance generalization.

\balance
\bibliographystyle{IEEEtran}
\bibliography{icjnnBib}
\end{document}